\documentclass[10pt,twocolumn,letterpaper]{article}

\usepackage{cvpr}
\usepackage{times}
\usepackage{epsfig}
\usepackage{graphicx}
\usepackage{amsmath}
\usepackage{amssymb}
\usepackage{algorithm}
\usepackage{algpseudocode}
\usepackage{xcolor}
\usepackage{multirow}

\begin{document}

\title{UFQA: Utility guided Fingerphoto Quality Assessment}

\author{Amol S. Joshi, Ali Dabouei, Jeremy Dawson, Nasser Nasrabadi\\
West Virginia University\\
Morgantown, WV, USA\\
{\tt\small asj00003@mix.wvu.edu}
}

\maketitle
\thispagestyle{empty}

\begin{abstract}
Quality assessment of fingerprints captured using digital cameras and smartphones, also called fingerphotos, is a challenging problem in biometric recognition systems. As contactless biometric modalities are gaining more attention, their reliability should also be improved. Many factors, such as illumination, image contrast, camera angle, etc., in fingerphoto acquisition introduce various types of distortion that may render the samples useless. Current quality estimation methods developed for fingerprints collected using contact-based sensors are inadequate for fingerphotos. We propose Utility guided Fingerphoto Quality Assessment (UFQA), a self-supervised dual encoder framework to learn meaningful feature representations to assess fingerphoto quality. A quality prediction model is trained to assess fingerphoto quality with additional supervision of quality maps. The quality metric is a predictor of the utility of fingerphotos in matching scenarios. Therefore, we use a holistic approach by including fingerphoto utility and local quality when labeling the training data. Experimental results verify that our approach performs better than the widely used fingerprint quality metric NFIQ2.2 and state-of-the-art image quality assessment algorithms on multiple publicly available fingerphoto datasets.
\end{abstract}

\section{Introduction}
Quality estimation in biometric recognition systems is crucial for several reasons. First, acquisition systems must ensure the quality of biometrics during enrollment. Allowing poor quality samples in the datasets will lead to inefficient use of resources. Second, having a majority of high-quality samples in the reference set will allow the recognition systems to match the subject with lower false non-match rates (FNMR) accurately. Finally, poor quality samples may introduce spurious features and fool the matcher into incorrectly recognizing the subject. This behavior poses serious security threats in sensitive areas such as law enforcement and access control. Therefore, the objective of the quality assessment task is to identify the recognition utility of all samples in the corpus, thereby reducing the false match rate (FMR).
\begin{figure}
\begin{center}
\includegraphics[width=0.8\linewidth]{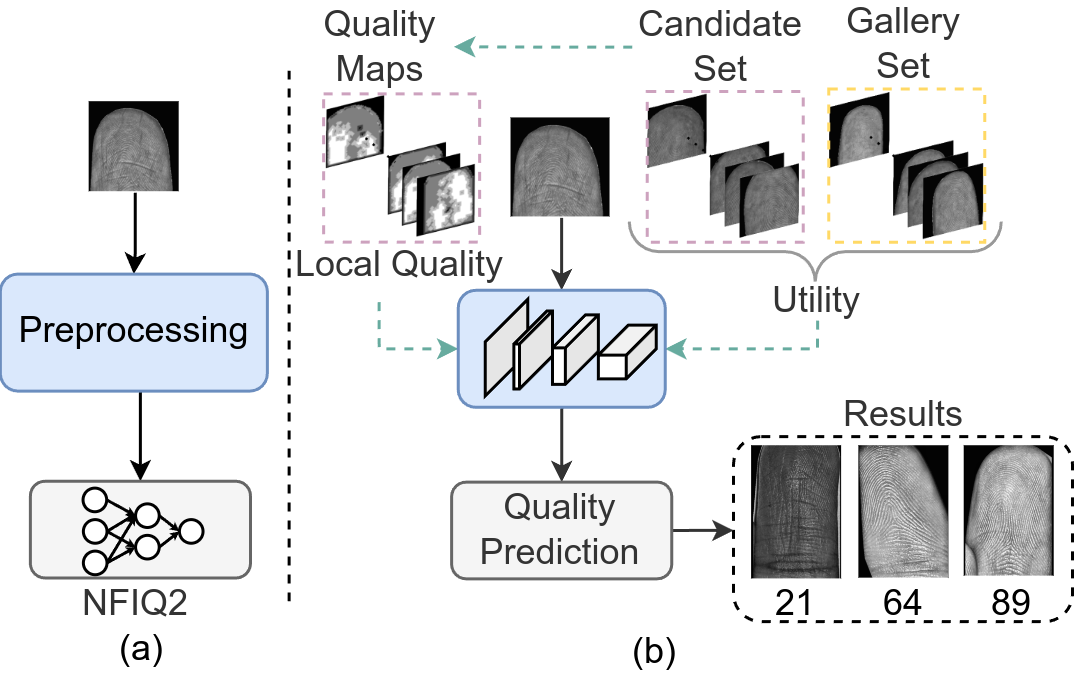}
\end{center}
   \caption{Comparison between fingerphoto quality assessment methods. (a) Conventional approach to obtain quality through pre-processing, such as filtering, histogram equalization, etc. and applying NFIQ2. (b) Our proposed UFQA model that incorporates fingerphoto utility and local information to predict quality.}
\label{fig:intro}
\end{figure}

In recent years, contactless biometric modalities have become a preferred choice for personal identification. The acquisition of such biometrics is hygienic and convenient for the user. However, new challenges arise with new acquisition systems. In terms of fingerprint modality, fingerprints are collected using conventional touch-based sensors. Whereas, contactless acquisition systems use smartphones, digital cameras, and touchless sensors \cite{chowdhury2022contactless} to capture fingerprints, which we refer to as fingerphotos. Fingerprint acquisition can be affected by a variety of factors, such as skin conditions, moisture, dryness, etc. In addition to this, depending on user cooperation, the interaction between the sensor surface and the finger causes the fingerprints to contain elastic distortions and inconsistent ridge-valley patterns.  

On the other hand, fingerphoto acquisition is less constrained than contact-based methods. As a result, fingerphotos suffer from various drawbacks. First, the cylindrical shape of the finger causes perspective distortions, which are responsible for a significant loss of peripheral areas of the finger. Second, camera positioning and lighting conditions introduce inconsistent illumination and image contrast, resulting in poor ridge-valley patterns. Third, if the subject has trembling fingers, the fingerphotos will turn out blurry and challenging to match. Additionally, because of all these arbitrary covariate (nuisance) factors, the fingerphotos exhibit regional distortions. A poorly focused camera may capture a particular image region, and improper illumination may occlude visible ridges. Lastly, the fingerphotos are not readily usable for the current recognition systems. A pre-processing pipeline is required to extract useful features that can be used for matching \cite{birajadar2019towards, lin2018matching, 6313540}. 

Therefore, the fingerphoto quality metric should account for the shortcomings mentioned above. This can be achieved through robust feature extraction that analyzes the fingerphoto characteristics with respect to useful identity information. Hand-crafted feature extraction has been around for decades. Initially, NIST Fingerprint Image Quality 2 (NFIQ2.2) \cite{539646} proposed extracting several feature maps, such as a direction map, low contrast, low flow, high curve, and a summarized quality map to estimate quality. These features are highly effective in predicting the utility of the fingerprint; however, only some of the features are relevant to the case of fingerphotos. The authors in \cite{6712750} considered three factors, ridge quality, minutiae reliability, and finger position, to compute various features for latent fingerprint quality. The same concept of feature extraction is proposed for fingerphotos that explicitly account for camera settings and lighting conditions of the images during acquisition \cite{labati2010neural, li2013quality, 6313540}.  

Although the methods that use hand-crafted feature extraction are reliable, computational expense is still in question. Quality assessment must be equally efficient for faster and less constrained acquisition systems. Olsen \textit{et al.} showed a promising direction to reduce the computational complexity of conventional methods using the neural network-based algorithm \cite{aastrup2013self}. Furthermore, recent advances in representation learning have increased the capabilities to extract reliable features relevant to desired modalities \cite{nguyen2018robust, nguyen2020universal, soleymani2021quality}. Many approaches using deep neural networks have been proposed for biometric recognition tasks \cite{minaee2023biometrics}. Therefore, a representation-learning-based approach for fingerprint quality estimation appears to be a promising candidate. 

Nonetheless, there are several challenges to estimate the quality of finger photographs. In most cases, the border areas of a fingerphoto are either the background or the distorted ridges because of the curved shape of the finger. Uneven ridge-valley patterns and non-uniform distortions affect the visibility of region of interest (ROI). To deal with the regional distortions, we propose an additional task in our quality estimation model to learn the features of the regional quality of the fingerphoto. In addition, the lack of availability of large fingerphoto training datasets undermines the learning capacity of a neural network. Therefore, we use existing large contact-based fingerprint datasets for transfer learning. It ensures that the network learns to extract reliable fingerprint representations. Then, we use the available fingerphoto datasets to fine-tune this model. Data annotation using reliable labels is another challenge of the quality estimation system. We use a labeling algorithm developed for fingerprints and modify it to satisfy our constraints. Lastly, a recognition system is essential to evaluate the estimated quality. To the best of our knowledge, current recognition systems are not entirely optimized for matching fingerphotos. Therefore, we have to rely on these systems to evaluate our quality metric. We provide a general overview of the conventional approach to fingerphoto quality assessment and our proposed UFQA in Figure \ref{fig:intro}. For the sake of brevity, we use \lq fingerprint\rq~and \lq fingerphoto\rq~terminologies for the finger samples collected from a touch-based sensor and a smartphone/digital camera, respectively.

Our main contributions are listed below:
\begin{itemize}
    \item We propose UFQA, a self-supervised dual encoder framework with a fusion module and a quality prediction network to assess fingerphoto quality that operates as a predictor of recognition utility.
    \item We make a holistic labeling process to consider image characteristics and identity information of fingerphotos for training the model.
    \item We evaluate our method on multiple publicly available datasets and demonstrate its effectiveness by comparing it with three image quality assessment (IQA) algorithms.
\end{itemize}

This paper is organized as follows: In Section \ref{sec2}, we review the related work and the literature. Next, in Section \ref{sec3}, we discuss the label generation process and the proposed quality assessment model. Section \ref{sec4} addresses datasets and evaluation experiments to analyze the performance of our quality metric. Later in Section \ref{sec5}, we discuss the ablation study to evaluate some of the critical components of the proposed method. Finally, Section \ref{sec6} summarizes the work and concludes the paper.

\section{Related Work}\label{sec2}
In this section, we review previous work on the assessment of the quality of fingerprints and fingerphotos.
\subsection{Fingerprint Quality Assessment}
Many approaches have been proposed to estimate the quality of fingerprints. These methods are based on local and global hand-crafted feature extraction \cite{lee2005model,lim2002fingerprint, shen2001quality, tabassi2005novel, xie2010effective}. Alonso \textit{et al.} show a comparative analysis of the various fingerprint quality assessment methods \cite{alonso2007comparative}. Fronthaler \textit{et al.} use the orientation tensor of the fingerprint to assess the signal strength in the fingerprint image. In addition, they fuse multiple algorithms to obtain the quality score \cite{fronthaler2008fingerprint}. Yao \textit{et al.} proposes a weighted sum of features extracted from fingerprints. They use prior, texture-based, and minutiae-based features to compute quality \cite{yao2015fingerprint}. NIST NFIQ2 is a widely used fingerprint quality assessment algorithm that has high predictive power for the recognition utility of the sample. It uses a vector of 24 features to compute the quality score \cite{918416}. These features include frequency domain analysis, orientation certainty, ridge valley uniformity, minutiae quality, ROI-based features, and local qualities. Recent approaches use neural network-based feature extraction to estimate quality. Olsen \textit{et al.} proposed using self-organizing maps to cluster fingerprint patches based on spatial information. The high-level representation of the fingerprint is then fed to a Random Forest classifier \cite{aastrup2013self}. The authors in \cite{terhorst2021midecon} assess the confidence in the detection of minutiae and use it to compute the quality of the fingerprint.

\subsection{Fingerphoto Quality Assessment}
With increasing exposure to fingerphotos, multiple methods have been proposed to assess the quality of the image of fingerphotos. Labati \textit{et al.} extract 45 features from the fingerphoto and select the subset of the best features to feed to a neural network-based classifier \cite{labati2010neural}. Priesnitz \textit{et al.} propose preprocessing steps for fingerphotos such that NFIQ2 can be applied to estimate the quality \cite{priesnitz2020touchless}. The authors in \cite{li2013quality}  use a block-based approach to compute the quality of the entire image. They extract 12 features from image blocks and use a Support Vector Machine (SVM) \cite{hearst1998support} classifier to get a quality class. 

Multiple studies have been conducted to assess the fingerprint matching ability of fingerphotos \cite{birajadar2019towards, chowdhury2022contactless, 6313540}. Parziale and Chen propose a coherence-based approach that uses local gradients to estimate the quality of local patches \cite{parziale2009advanced}. Wild \textit{et al.}, while studying the interoperability between fingerprints and fingerphotos, proposes a fingerphoto enhancement process and uses NFIQ2 to estimate quality \cite{8739191}. Kauba \textit{et. al.} used edge detection to estimate ridges in the image and calculate quality based on a threshold \cite{s21072248}.

The review of the literature shows that the assessment of fingerphoto quality is still an open problem. It is essential to fine-tune the feature extraction and selection process to assess the fingerphoto utility. In addition, the regional distortions and the nature of fingerphotos differ significantly from the fingerprints. Therefore, the quality metric should account for these factors. The NFIQ2 is a robust tool for fingerprint quality estimation, which is a promising direction to address these issues.
\begin{figure}
\begin{center}
\includegraphics[width=0.65\linewidth]{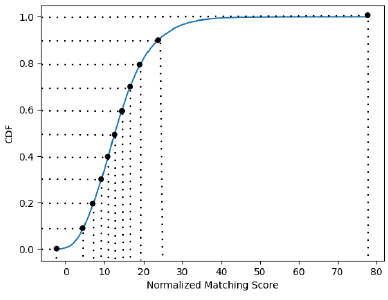}
\end{center}
   \caption{ECDF plot with normalized matching scores on the x-axis and CDF on the y-axis. The dots in the plot show the split location for each class.}
\label{fig:ecdf}
\end{figure}
\begin{figure*}
\begin{center}
\includegraphics[width=0.85\linewidth]{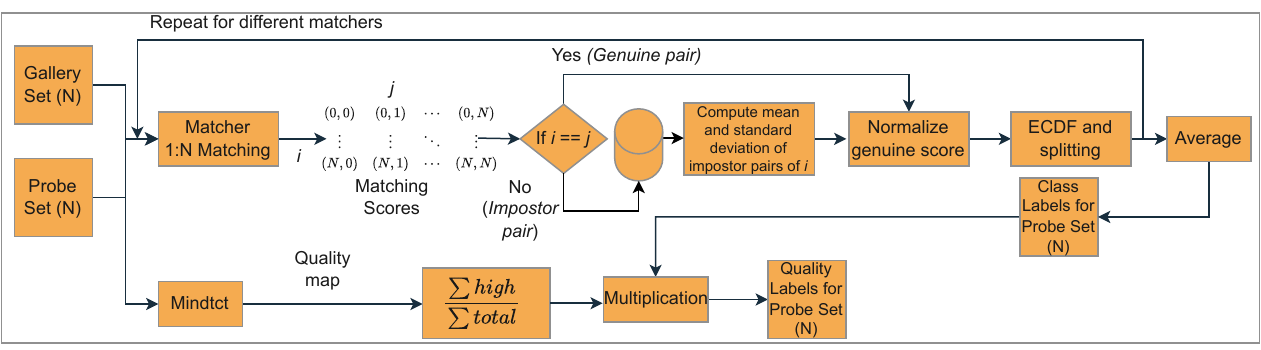}
\end{center}
   \caption{Quality label generation process flow. The labels obtained in the figure are for the probe set. We switch the gallery and probe sets and repeat the process to get the labels for the gallery set.}
\label{fig:flow_chart}
\end{figure*}
\section{Proposed Method}\label{sec3}
Our goal in this study is to assess the image quality based on the assumption that the low-quality images are difficult to match using any fingerprint matcher. To achieve this, we emphasize fingerphoto quality by learning better representations through matching scores of pairs of images. The proposed algorithm consists of two encoders to extract features, a feature fusion module, and a quality prediction network. The extracted features are sent to our proposed quality assessment module to return the quality of the fingerphoto. Here, we discuss the generation of quality labels in Section \ref{sec3.1}, the quality assessment model in Section \ref{sec3.2}, and finally, the implementation details in Section \ref{sec3.3}.
\subsection{Quality Label Generation}\label{sec3.1}
Labeling training data with a reliable quality score is essential in the quality assessment process using neural networks. Furthermore, in the case of biometric data, quality should be proportional to the recognition performance (i.e., utility) of the fingerprint. Therefore, we need a quality label generation process that considers crucial properties of the ridge patterns and minutiae points while deciding the fingerprint quality. 

Grother and Tabassi proposed a recognition performance-based procedure to annotate samples with the quality class \cite{grother2007performance}. The procedure involves using a fingerprint matcher to obtain similarity scores between the gallery and probe sets. Then, on the basis of the normalized scores, two sets are formed such that the pair belongs to one set if the matching score of the genuine pair is greater than all its impostor pairs. The two sets generate two empirical cumulative distribution functions (ECDFs) that are used to classify the reference samples into five categories. Finally, five fingerprint matchers are tasked to classify the same sample and assign the quality label to the sample if all matchers predicted the same class. If the assigned label is different, they modify the threshold and re-assign it. If disagreement persists, the fingerprint is discarded from the training set.

For the generation of fingerphoto labels, due to the limited amounts of publicly available data and the difference between fingerprints and fingerphotos, we make two modifications to the above procedure. First, we compute only one ECDF plot for the entire dataset for each matcher. To categorize the data, we empirically select ten classes, since having ten classes allows for more fine-grained quality scores. Figure \ref{fig:ecdf} shows the ECDF plot and the corresponding divisions. Second, fingerprint matchers play a vital role in the quality label generation process. The quality label might be affected depending on the features used to match two fingerprints. Therefore, bias toward a matcher is always possible if it is involved in the label generation process. As Grother and Tabassi suggested in \cite{grother2007performance}, we use two commercial matchers mentioned in Section \ref{sec4.2} to assign quality labels to mitigate this problem. In the event of disagreement on the label between the matchers, we adopt the policy of averaging the labels from the two matchers rather than discarding the samples.

\begin{figure}
\begin{center}
\includegraphics[width=0.7\linewidth]{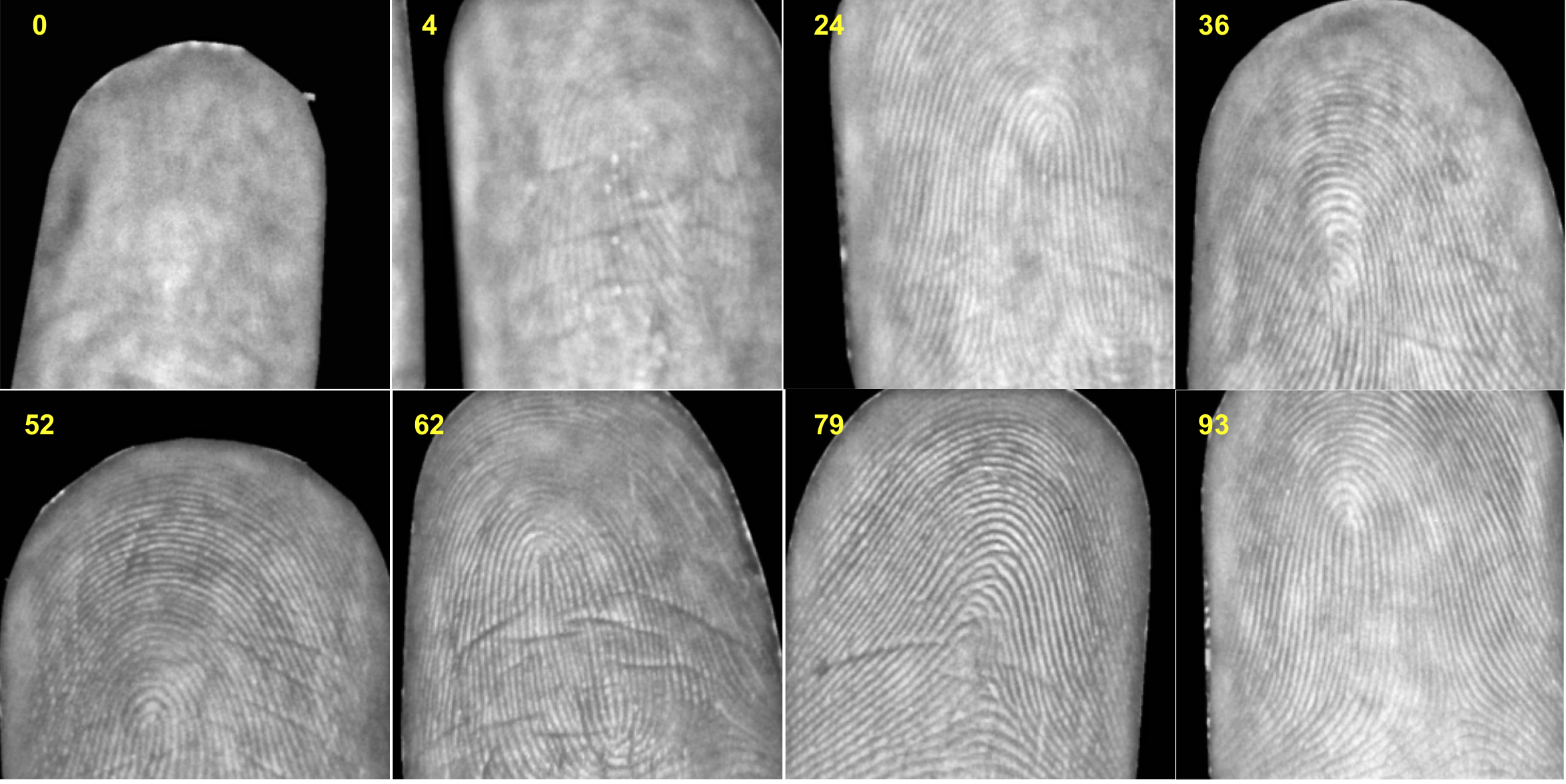}
\end{center}
   \caption{Generated quality labels for different samples. The number in the upper right corner represents the quality label for each image. Zoom in for better view.}
\label{fig:qual_label}
\end{figure}
\begin{figure*}
\begin{center}
\includegraphics[width=0.85\linewidth]{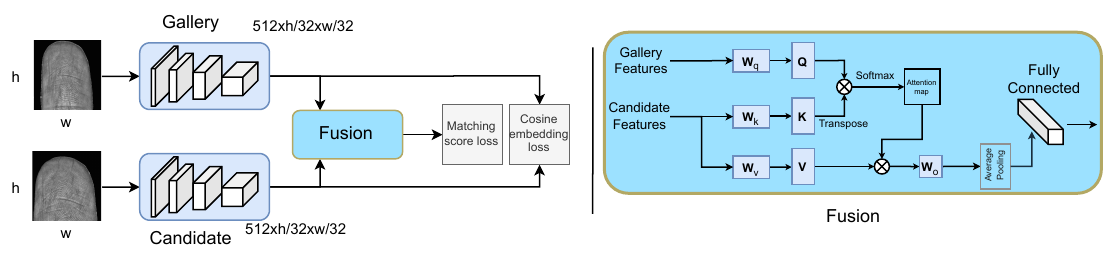}
\end{center}
   \caption{The architecture of the proposed UFQA model. The input to the network is two fingerphotos, and the output features from the candidate encoder are used to predict the image quality between $0$ and $100$ using the quality prediction network. The feature fusion module is displayed on the right side of the figure.}
\label{fig:network}
\end{figure*}
This process is well grounded to annotate fingerprints. However, fingerphotos differ in the stages of acquisition, processing, and feature extraction compared to fingerprints. A vital drawback in the method discussed above is the dependence on the feature extraction of the matchers involved in the generation of the labels. Failure of matchers to extract reliable features from poor quality fingerphotos may result in lower genuine scores and higher impostor scores. This causes an imbalance in the ECDF, resulting in mislabeled data. Therefore, to alleviate this problem, we introduce the local quality ratio of the fingerphoto to weigh the generated quality class. It prevents poor-quality images from having higher-quality labels, while preserving the labels of high-quality fingerphotos. The local quality ratio is the ratio between the number of high-quality patches and the total number of patches. The quality of patches is obtained using the Mindtct minutiae detection algorithm from NIST \cite{51496}. Each patch is assigned a quality between zero and four, with zero being poor quality. We set two as the threshold to count the patch as high-quality. The process of generating the label for each image is shown in Figure \ref{fig:flow_chart}. Figure \ref{fig:qual_label} provides some samples with the quality labels assigned using the algorithm provided.

\subsection{Quality Assessment Model}\label{sec3.2}
For brevity, we discuss the quality assessment model separately in two parts: feature extraction is discussed in \ref{sec:3.2.1} and quality prediction is discussed in \ref{sec:3.2.2}.
\subsubsection{Feature extraction}\label{sec:3.2.1}
Our proposed quality assessment model, UFQA, comprises a self-supervised learning framework with two encoders and a feature fusion module. Both encoders use a ResNet18 \cite{he2016deep} architecture without global average pooling and fully connected layers. In the data labeling process, we observed the effectiveness of including matching scores to assess fingerphoto quality. We exploit this further by proposing a self-supervised dual encoder scheme that receives a candidate and a gallery fingerphoto. Note that the gallery fingerphoto is not a high-quality reference image, as widely used in reference-based image quality assessment algorithms. In our setup, the input fingerphotos could be a mated or non-mated pair. The embeddings from the two encoders are fused together in the feature fusion module. 

This module consists of a self-attention block followed by a fully connected layer. The conventional self-attention mechanism from \cite{vaswani2017attention} is used to discover the relationship between the two embeddings. The self-attention serves the purpose of highlighting similarities in identity-related information from the two fingerphotos if they belong to the same subject. The self-attention module applies a fully connected layer to the features received from the candidate encoder to obtain key and value representations, while the query comes from the gallery features. The attention map is computed using a dot product between query and transposed key representations followed by a softmax layer. Finally, the dot product between the value and the attention map is computed and is again projected to return to the original shape. The network architecture for the feature extraction part is provided in Figure \ref{fig:network}. 

This way of fusing the two features helps the network to focus on minutiae points and ridge-valley patterns in the fingerphoto and it tries to learn the discrepancies in non-mated pairs. To this aim, we train the dual encoder and the fusion module jointly with the following objective function:
\begin{equation}
\begin{split}\label{obj_feat}
    \mathcal{L}_{feat} = \lambda_1\|s - F(\varphi(.), \omega(.))\|_{2} + \lambda_2\mathcal{L}_{sim},
\end{split}
\end{equation}
where, $\varphi(.), \omega(.)$ are candidate and gallery encoders, respectively. $F(.)$ is the fusion module and $s$ is the matching score for the corresponding pair. The matching score for the input pair is obtained from two matchers that use minutiae-based fingerprint matching. $\lambda_1$ and $\lambda_2$ are the scaling coefficients set to $10.0$ and $2.0$, respectively. Lastly, $\mathcal{L}_{sim}$ is the cosine embedding loss as given below:
\begin{equation}
\begin{split}
\mathcal{L}_{sim} = \begin{cases} 1 - cos(\boldsymbol{x_c}, \boldsymbol{x_g}) & \text{if $y$ = $1$}\\
\max(0, cos(\boldsymbol{x_c}, \boldsymbol{x_g})- m)) & \text{if $y$ = $-1$}.
\end{cases}
\end{split}
\end{equation}
Here, $\boldsymbol{x_c}$ and $\boldsymbol{x_g}$ are embeddings produced by the encoders $\varphi$ and $\omega$, respectively. $y$ is the label indicating the mated or non-mated pair, while $m$ is the margin set to 0. $cos(.)$ is the cosine similarity given by:
\begin{equation}
cos(\boldsymbol{x_c}, \boldsymbol{x_g}) = \frac{\boldsymbol{x_c} \cdot \boldsymbol{x_g}}{\max(\|\boldsymbol{x_c}\|_2 \cdot \|\boldsymbol{x_g}\|_2, \epsilon)},
\end{equation}
where, $\epsilon$ is set to $1e-8$. This objective function minimizes the distance between two embeddings if they belong to the same subject. Moreover, this injects the utility of the candidate fingerphoto into quality assessment via predicting the matching score and bringing similar embeddings closer. Finally, during inference, only the candidate encoder $\varphi$ is used to get the features for quality prediction.
\subsubsection{Quality prediction}\label{sec:3.2.2}
The quality prediction network $P(\varphi(.))$ learns to estimate quality using the embeddings of the candidate fingerphoto. The embeddings acquired from the candidate encoder are rich with the necessary information from the fingerphoto. Therefore, the same information used for the matching can be wielded to predict a quality score. To do this, we use an average pooling layer and a fully connected layer followed by a sigmoid function to obtain the quality score between $0$ and $100$. 

Fingerphotos often suffer from distortions due to illumination, lightning conditions, self-occlusion, etc. Such factors affect the minutiae and ridge patterns non-uniformly. Therefore, accounting for local distortions is imperative in computing the overall image quality \cite{ezeobiejesi2018latent, terhorst2021midecon}. To this end, we use regional quality information using $C(\varphi(.))$ to improve quality prediction. Here, $C$ is a multi-layer perceptron (MLP) with three fully connected layers and Leaky ReLU as activation. We use the embeddings from the candidate encoder $\varphi(.)$ to learn the regional quality distribution in the fingerphoto. This additional learning works as an auxiliary task that benefits the primary task of global quality estimation. We use the quality maps from Mindtct to train $C(.)$. More analysis on this is discussed in Section \ref{sec5.2}. 

Our primary task is to estimate a single score for the entire image. However, quality maps include spatial information on quality. Therefore, in addition to the L2 loss in pixel space, we incorporate error minimization of the mean and standard deviation of the quality map. This ensures that the overall distribution of the predicted quality map remains closer to the ground truth. The objective function for training the quality prediction network is as follows:
\begin{equation}
\begin{split}\label{obj_quality}
    \mathcal{L}_{qual} &= \lambda_3\| q_g - F(\varphi(I_c))\|_{2}\\ &+ \lambda_4(\| \boldsymbol{q_r} - C(\varphi(I_c))\|_{2}\\ &+ \|\mu(\boldsymbol{q_r}) - \mu(C(\varphi(I_c)))\|_{2}\\ &+ \|\sigma(\boldsymbol{q_r}) - \sigma(C(\varphi(I_c))) \|_{2}),
\end{split}
\end{equation}
where, $I_c$ represents the input fingerphoto. $q_g$ and $\boldsymbol{q_r}$ represent the ground truth quality score and the quality map, respectively. $\lambda_3$ and $\lambda_4$ are scaling factors and are empirically set to $0.1$ and $10$, respectively.
\subsection{Implementation Details}\label{sec3.3}
We initialize the network with the ResNet-18 architecture pre-trained on ImageNet \cite{5206848} and train the quality predictor $P(\varphi(.))$ on fingerprints. Since fingerphotos are not as abundant as fingerprints, we train the model to learn the quality estimation of fingerprints and transfer the knowledge to fingerphotos. For training the model with fingerprints, we use NFIQ2.2 to generate quality labels and minimize L2 loss. We assume that, by using a sufficiently large dataset, the network can learn to extract features relevant to NFIQ2.2. Further, to modify the network parameters to suit the fingerphoto quality prediction task, we fine-tune it using the labels generated from the approach discussed in Section \ref{sec3.1}.

For model training, we used the PyTorch framework \cite{NEURIPS2019_9015} with two Nvidia RTX 3090 Ti GPUs. Adam optimizer \cite{kingma2014adam} was used to train the model with a learning rate of $2e-4$ and a weight decay of $5e-4$. A polynomial scheduler was used to decay the learning rate. We trained the network on fingerprints for 50 epochs and later fine-tuned it for 30 epochs for fingerphotos with a batch size of 100.
\begin{table}
\begin{center}
\resizebox{\columnwidth}{!}{%
\begin{tabular}{|l|l|c|c|c|}
\hline
Matcher & Method & PolyU & RidgeBase & Multimodal \\
\hline\hline
\multirow{4}{*}{Bozorth3} & NFIQ2.2 & $0.0552$ & $0.1826$ & $0.0730$\\
& HyperIQA & $0.0558$ & $0.1816$ & $0.0760$\\
& AIT & $0.0576$ & $\textbf{0.1808}$ & $0.0740$\\
\cline{2-5}
& UFQA (Ours) & $\textbf{0.0549}$ & $0.1810$ & $\textbf{0.0675}$\\
\hline
\multirow{4}{*}{VeriFinger SDK} & NFIQ2.2 & $0.0019$ & $0.1557$ & $0.0365$\\
& HyperIQA & $0.0020$ & $0.1537$ & $0.0412$\\
& AIT & $0.0019$ & $0.1505$ & $0.0380$\\
\cline{2-5}
& UFQA (Ours) & $\textbf{0.0014}$ & $\textbf{0.1495}$ & $\textbf{0.0327}$\\
\hline
\multirow{4}{*}{IDKit SDK} & NFIQ2.2 & $0.0108$ & $0.1483$ & $0.0025$\\
& HyperIQA & $0.0111$ & $0.1478$ & $0.0024$\\
& AIT & $0.0101$ & $0.1430$ & $0.0022$\\
\cline{2-5}
& UFQA (Ours) & $\textbf{0.0092}$ & $\textbf{0.1427}$ & $\textbf{0.0013}$\\
\hline
\end{tabular}
}
\end{center}
\caption{pAUC achieved by the EDC plots of UFQA (Ours) and other image quality assessment algorithms. Our method outperforms the other methods for three different datasets.}
\label{tab:auc_table}
\end{table}
\begin{figure*}
\begin{center}
\includegraphics[width=0.75\linewidth]{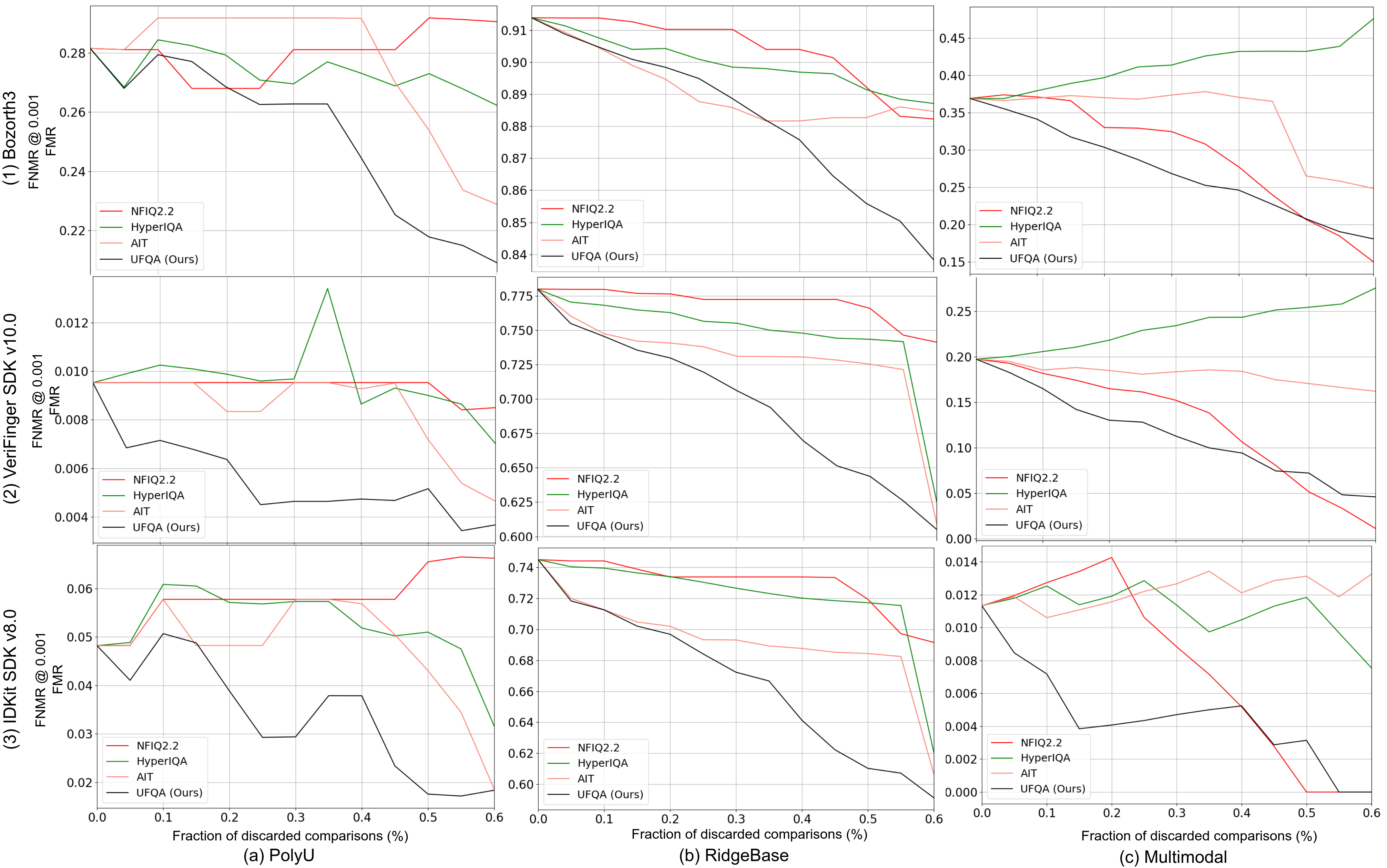}
\end{center}
   \caption{EDC plots for three different datasets using two COTS matchers. The columns represent the data sets, and the rows represent the matchers used in the evaluation experiment. All experiments were carried out at a fixed false match rate (FMR) of 0.001. In each plot, the red curve represents NFIQ2.2, green represents HyperIQA, pink represents AIT, while the black represents UFQA (Ours).} 
\label{fig:results}
\end{figure*}
\section{Experiments}\label{sec4}
In this section, we dive into the experiments we conducted to verify the usefulness of UFQA and compare it with state-of-the-arts in fingerphoto quality assessment. First, we discuss the datasets used for training and testing. Next, we discuss the evaluation protocol used to assess the performance of quality estimation during our experiments. Finally, we present the results obtained during the experiments.
\subsection{Datasets}\label{sec4.1}
As mentioned in Section \ref{sec3.3}, we first train our model on fingerprints and then fine-tune it on fingerphotos. For the first step, we used the Multimodal dataset \cite{wvu_dataset} and the NIST SD 302 dataset \cite{249671} to train and test the model on fingerprints. The Multimodal dataset has approximately 70,000 fingerprints from 2,175 subjects collected using two different sensors over a four-year period. Similarly, the NIST SD 302 dataset contains 7,701 samples from 200 subjects. We combine the fingerprints from the two datasets and apply six different augmentations: blurring, flipping, rotation, random cropping, adding Gaussian noise, and dropout. Using these augmentations, we generated around 0.5 million samples for training. Then, we use NFIQ2.2 to obtain the quality label for each fingerprint. 

In the second step, to fine-tune the model on fingerphotos, we use the Multimodal dataset and the RidgeBase dataset \cite{9648393, jawade2022ridgebase}. After combining both datasets, we get 28,545 fingerphotos from 688 subjects for training. To include samples of various qualities and make the model effective during image acquisition, we preserve the natural distribution of the datasets. Therefore, apart from skin masking and grayscaling, no preprocessing is performed on the datasets. As mentioned in Section \ref{sec3.1}, we require gallery and probe sets to obtain labels and train the model. To this end, we create genuine pairs using multiple instances of a fingerphoto from the same subject. The impostor pairs are created with a different fingerphoto from a different subject.

We use three publicly available datasets to evaluate the model trained on fingerphotos. The first evaluation set is contactless 2D to contact-based 2D fingerprint images database version 1.0 by the Hong Kong Polytechnic University \cite{lin2018matching} referred to as the PolyU dataset. This dataset contains 1,800 fingerphotos from 300 subjects collected using a digital CMOS camera. The second set has 2,229 samples from the RidgeBase dataset. Finally, the last dataset consists of 309 samples from the Multimodal dataset.
\subsection{Evaluation Protocol}\label{sec4.2}
Our quality assessment method assumes that quality is an indicator of the accuracy of fingerphoto recognition. Therefore, we use the EDC plots as evaluation criteria to assess the effectiveness of the predicted quality score \cite{grother2007performance}. The EDC is a performance metric used to evaluate the utility of quality assessment algorithms. It shows how the error rate changes as some low-quality samples are discarded. As low-quality samples are rejected from matching experiments, the error rate approaches zero. This behavior indicates that a good quality assessment algorithm should have low error and rejection rates. For plotting these curves, we used three matching algorithms. They are the open-source Bozorth3 matcher \cite{51496} and the COTS matchers VeriFinger SDK v10.0 by Neurotechnology \cite{verifinger} and IDKit SDK v8.0 by Innovatrics \cite{innovatrics}. We consider an FNMR with fixed FMR of $1e-3$ to report the verification error rates in the EDC plots. In addition, we compare the partial area under the curve (pAUC) for the EDC plots to quantify the error in a discard fraction range of $[0 - 0.2]$.

Furthermore, it is essential to analyze whether the model generates relevant embeddings for quality assessment. Therefore, we study the distribution of features using t-distributed Stochastic Neighbor Embedding (t-SNE) \cite{van2008visualizing}. t-SNE uses dimensionality reduction to visualize and interpret high-dimensional embeddings.
\begin{figure}
\begin{center}
\includegraphics[width=0.6\linewidth]{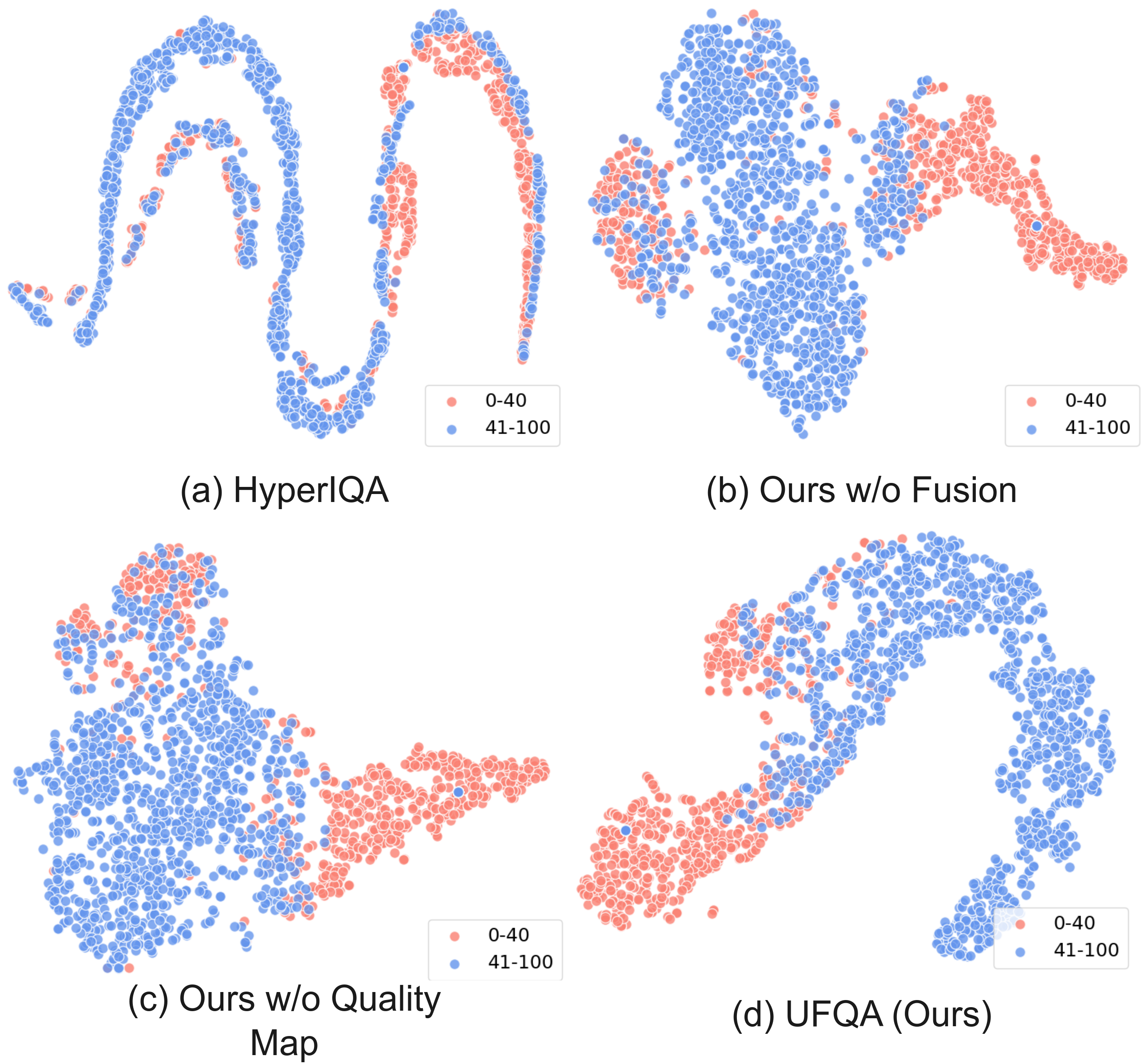}
\end{center}
   \caption{Comparison of 2D t-SNE visualization of features of samples from Multimodal dataset between (a) HyperIQA, (b) our model without the fusion module, (c) our model without the quality map supervision, and (d) the proposed model. The red and blue datapoints indicate low and high quality, respectively.}
\label{fig:tsne}
\end{figure}
\subsection{Results}\label{sec4.3}
Here, we discuss the results of the evaluation experiments on three test sets. The EDC plots obtained using three matchers are shown in Figure \ref{fig:results}. We compare our method with HyperIQA \cite{Su_2020_CVPR}, a sharpness-based metric proposed by Kauba \textit{et al.} \cite{s21072248} referred to as AIT, and the widely used quality estimation method NFIQ2.2. Previous studies have shown the applicability of NFIQ2.2 to estimate fingerphoto quality. Although trained on fingerprints, the NFIQ2.2 feature vector includes relevant features important for assessing fingerphoto quality \cite{jawade2022ridgebase, priesnitz2020touchless, 8739191}. However, pre-processing of fingerphotos is a crucial factor in quality assessment using NFIQ2.2. From the EDC plots, it is evident that overall our method outperforms other methods on all datasets using all three matchers. These results can be interpreted as the proposed network learning to extract relevant features inherently, which perhaps NFIQ2.2 and other methods fail to achieve. pAUC scores for the EDC plots are reported in Table \ref{tab:auc_table}. 

\begin{figure}
\begin{center}
\includegraphics[width=0.7\linewidth]{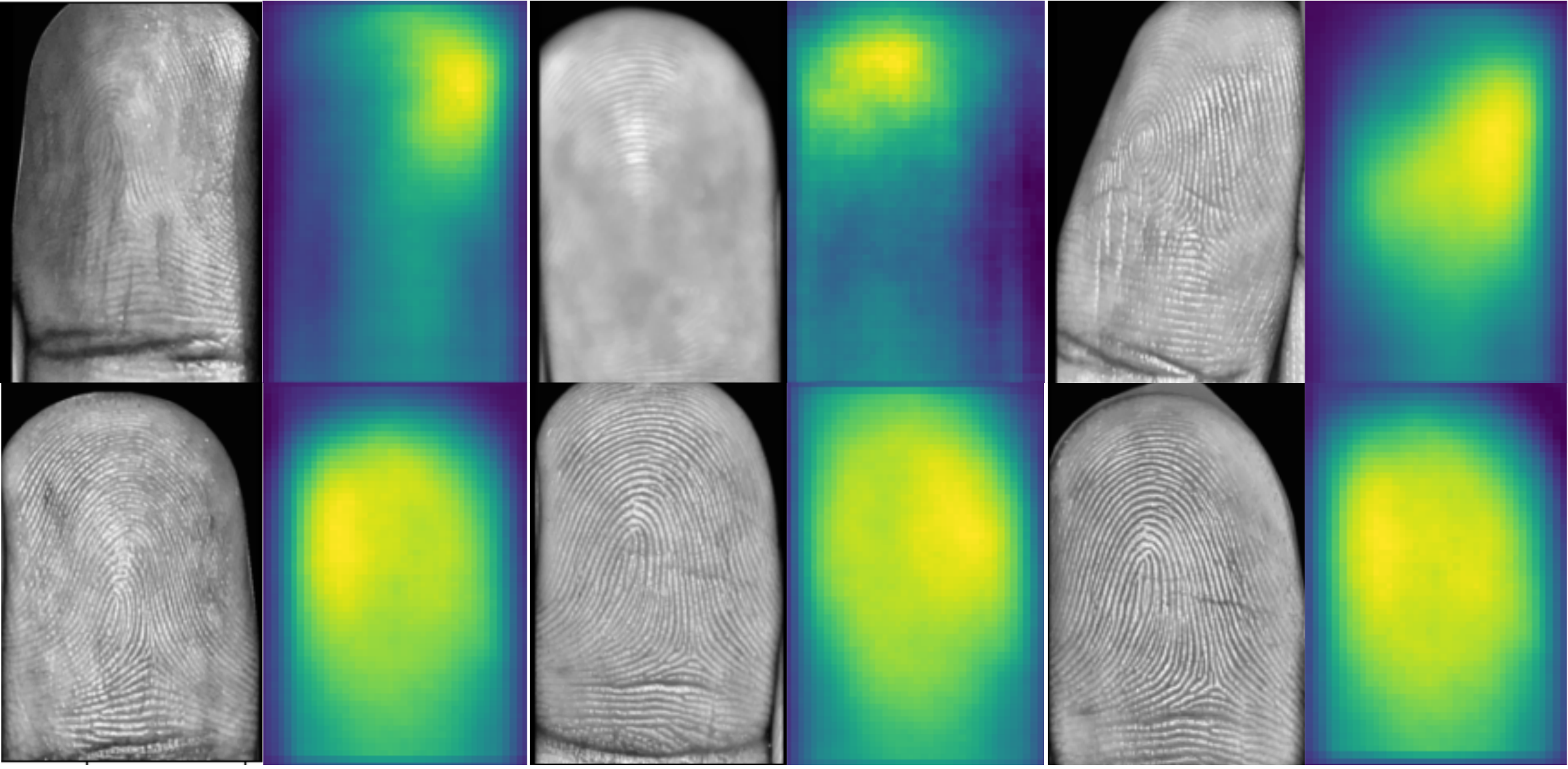}
\end{center}
   \caption{Samples of the quality map obtained using our method for six fingerphotos. The upper row shows low-quality samples with local distortions in the images, and the bottom row shows high quality samples.}
\label{fig:regional}
\end{figure}

In addition to the EDC plots, Figure \ref{fig:tsne} shows the t-SNE visualization of the features extracted from our model and HyperIQA. It is evident that the proposed model distinctly differentiates between low- and high-quality fingerphotos. For visualization, samples with quality scores less than 40 are considered low-quality, and the remaining are marked as high-quality.
Lastly, we visualize the regional quality maps to illustrate the ability of the network to capture semantic details such as black background, distorted regions, high-quality ridge-valley patterns, etc. Figure \ref{fig:regional} shows several fingerphotos and their corresponding predicted quality maps. Due to the self-attention mechanism and the weakly supervised regional quality estimation approach, the network is capable of focusing on the useful regions and disregarding the regional distortions that affect the overall quality.

\section{Ablation Study}\label{sec5}
In this section, we analyze two aspects of the quality assessment process. First, we discuss the effectiveness of an additional encoder to obtain better representations of the fingerphotos. Later, Section \ref{sec5.2} addresses the improvement in quality assessment performance by adding an auxiliary task to predict quality maps.
\subsection{Dual Encoder for Better Representations}\label{sec5.1}
Learning better representations is vital in our approach to IQA. To this aim, we introduced the dual encoder scheme, which accepts a pair of gallery and probe fingerphotos. Initially, we train the model with a single encoder without the fusion module. However, since it focuses more on the image characteristics than the matching ability of the fingerphoto, we observed that features extracted from this model possess a limited capability to represent various quality levels. Figure \ref{fig:tsne} (b) shows that the data points that represent low- and high-quality fingerphotos are extensively intermixed. On the contrary, our proposed model with the self-supervised dual encoder framework allows meaningful representations with the help of matching scores and cosine embedding loss. This is reflected in Figure \ref{fig:tsne} (d) and in the EDC plot in Figure \ref{fig:abl_regional}.
\subsection{Effect of Regional Quality Estimation}\label{sec5.2}
\begin{figure}
\begin{center}
\includegraphics[width=0.75\linewidth]{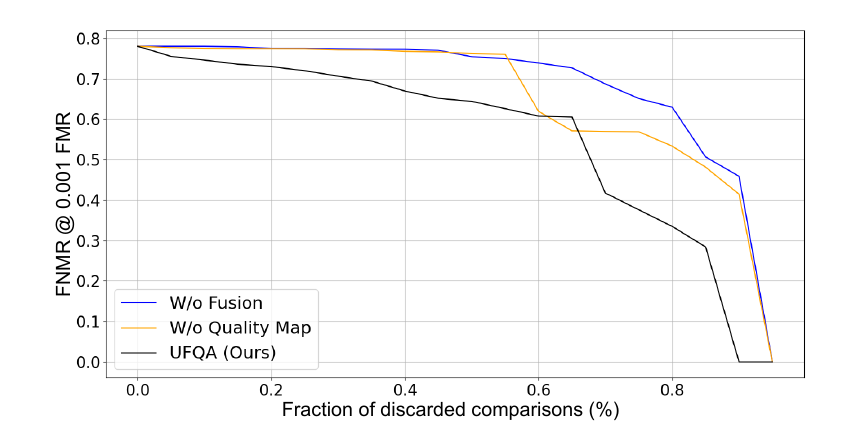}
\end{center}
   \caption{EDC plot for the model without the regional quality estimation, shown by the red curve, and the proposed model, shown by the green curve. VeriFinger SDK v10.0 was used to obtain the matching scores.}%
\label{fig:abl_regional}
\end{figure}

More high-quality patches in the image advocate that matchers can extract better features from the image. This ultimately leads to a reliable quality label obtained through normalized matching scores. Here, we attempt to test this hypothesis by eliminating the additional task that predicts the quality map. Figure \ref{fig:abl_regional} shows the EDC plot for the model without the regional quality estimator and the proposed model. We can notice a considerable drop in quality prediction performance. Furthermore, the comparison in Figures \ref{fig:tsne} (c) and (d) supports the hypothesis that the addition of the quality map helps to generalize the network and create a sufficient distinction between low- and high-quality fingerphotos. 
\section{Conclusion}\label{sec6}
In this study, we presented, UFQA, an approach to estimate fingerphoto quality that accounts for the image utility as well as local quality. We proposed a self-supervised dual encoder-based architecture to predict a fingerphoto quality score. Moreover, we used a secondary encoder to obtain representations relevant to the identity of the fingerphotos. We pose the quality metric as an indicator of recognition performance to generate the labels and evaluate the model. Our evaluation of various publicly available datasets shows the efficacy and robustness of UFQA compared to current fingerprint quality estimators. We analyze various aspects of our quality assessment model in an extensive ablation study. The quality assessment of fingerphotos is a vital problem with the increasing use of contactless fingerprints, and we hope that this work contributes to the advancement of fingerphoto biometric.

\noindent{\textbf{Acknowledgements.}} This material is based upon a work supported by the Center for Identification Technology Research and the National Science Foundation under Grant \#1650474.

{\small
\bibliographystyle{ieee}
\bibliography{egbib}
}

\end{document}